\documentclass{article}

\usepackage[margin=2.0cm]{geometry}

\usepackage{amsmath}
\usepackage{amsfonts}
\usepackage{amssymb,latexsym}
\usepackage{amsthm}
\usepackage{bm}

\usepackage{graphicx}
\usepackage{tabularx}
\usepackage{adjustbox}
\usepackage{siunitx}
\usepackage{subcaption}
\usepackage{rotating}
\usepackage{multirow}
\usepackage{booktabs}
\usepackage{float}

\usepackage[english]{babel}
\usepackage[utf8]{inputenc}

\usepackage{algorithm}
\usepackage[noend]{algpseudocode}

\usepackage{xcolor}
\usepackage{hyperref}

\providecommand{\emailauthor}[2]{}
\providecommand{\urlauthor}[2]{}
\providecommand{\corref}[1]{}
\providecommand{\cortext}[2][]{}
\providecommand{\ead}[1]{}
\providecommand{\address}[2][]{}

\title{Adaptive RBF-KAN: A Comparative Evaluation of Dynamic Shape Parameters in Kolmogorov-Arnold Networks}

\author{
Roberto Cavoretto$^{1,2}$,
Alessandra De Rossi$^{1,2}$,
Adeeba Haider$^{1,2,*}$,
Amir Noorizadegan$^{3}$ \\[4pt]
{\small $^{1}$Department of Mathematics ``Giuseppe Peano'', University of Torino, via Carlo Alberto 10, 10123 Torino, Italy} \\
{\small $^{2}$Member of the INdAM Research Group GNCS} \\
{\small $^{3}$Department of Mathematics, Hong Kong Baptist University, Hong Kong SAR, China} \\[4pt]
{\small $^{*}$Corresponding author: \texttt{adeeba.haider@unito.it}}
}

\date{}

\begin{document}

\maketitle

\noindent
\begin{abstract}
\noindent
Kolmogorov-Arnold Networks (KANs) approximate multivariate functions using learnable univariate edge functions, typically parameterized by B-spline bases. Although effective, spline-based implementations can be computationally expensive. A modified version of KANs, called FastKAN, improves efficiency by replacing splines with Gaussian radial basis functions (RBFs), but it relies on a fixed kernel and shape parameter. In this work, we extend the RBF-based KAN framework by introducing a broader family of radial basis kernels and by initializing the kernel shape parameter using leave-one-out cross-validation (LOOCV). To the best of our knowledge, this is the first study that integrates LOOCV-based kernel scale estimation with deep KAN training. We also introduce Mat\'ern and Wendland kernels into the KAN framework for the first time, enabling more flexible basis representations beyond the Gaussian kernel used in FastKAN. The LOOCV estimate provides a data-driven initialization of the kernel scale, which is subsequently refined during network training. The proposed adaptive RBF-KAN is evaluated on several two-dimensional benchmark functions. The results highlight the importance of kernel selection and adaptive shape parameters, with different kernels showing advantages for smooth functions, discontinuities, and oscillatory patterns. Overall, combining LOOCV-based initialization with adaptive kernel learning provides a practical strategy for improving RBF-based KAN models.
\end{abstract}

\vspace{0.5em}



\section{Introduction} \label{sec:1}
The approximation of multivariate functions using scattered data interpolation is considered as an important problem in community of researchers working in scientific computing and machine learning. From simple functions it ranges applications from pattern recognition and data mining to weather prediction and image analysis. 

To explicitly study these problems, a huge variety of methods and practical approaches could be found in literature, including radial basis function networks (RBFNs) \cite{mai_duy03}, kernel-based learning methods \cite{baudat01}, and artificial neural networks (ANNs) \cite{devore21}. For approximation of functions, regression modeling, image classification tasks \cite{pinkus99} and many other related applications, multilayer perceptrons (MLPs) are considered as a standard architecture from decades in the field of neural networks.

MLPs consist of stacked layers of neurons connected through linear transformations followed by nonlinear activation functions. 
Their theoretical foundation is provided by the universal approximation theorem \cite{hornik89}, which states that sufficiently wide networks can approximate arbitrary continuous functions. 
This capability has made MLPs widely used in supervised learning tasks such as regression and classification, typically trained using gradient-based backpropagation methods \cite{rumelhart86, Amir24a,Amir24b,Amir26_reg}. 
Modern deep architectures further extend their applicability to complex tasks including image classification and large-scale data analysis \cite{coskun03}. 

Despite their flexibility, MLPs also face several well-known limitations. 
Because the nonlinearities are fixed activation functions applied at the nodes, the model has limited ability to adapt its functional representation to the structure of the underlying data. 
As a result, achieving high approximation accuracy may require large network sizes and significant parameter counts. 
These limitations have motivated the exploration of alternative neural architectures in which the nonlinear components of the network are themselves learned from data.

Kolmogorov-Arnold Networks (KANs), introduced by Liu et al.~\cite{liu24}, represent one such alternative architecture. 
KANs are \textit{inspired} by the Kolmogorov-Arnold (KA) representation theorem \cite{kolmogorov56}, which states that multivariate continuous functions can be represented as sums of compositions of univariate functions. 
In contrast to MLPs, where nonlinearities are applied at the nodes, KANs place learnable univariate functions on the edges connecting neurons \cite{Amir26}. This modification in structure of the architecture allows the network to learn flexible representations directly from data since it replaces fixed activation functions with adaptive functional components. Moreover, it also provides notable improvements in terms of interpretability.

In the original implementation of KAN proposed in \cite{liu24}, the edge functions are parameterized using B-spline basis functions. Although, on the one hand, this representation offers strong approximation capability and local control, on the other hand, the spline-based formulation can introduce a significant increase in computational cost during training. 

To address this limitation, several alternative KAN architectures have been proposed. For example, FastKAN replaces spline bases with Gaussian radial basis functions (RBFs) to accelerate computation \cite{li24}, while EfficientKAN reformulates the implementation to improve memory efficiency and training speed \cite{blealtan24}. 
Other variants explore alternative basis representations, including Fourier-based KAN architectures \cite{Zhang2025} and polynomial-based approaches such as Gottlieb-KAN \cite{Seydi2024}. 
These developments highlight the modular nature of the KAN framework, where different basis functions can be incorporated to balance accuracy, computational efficiency, and representational flexibility \cite{Amir26}.

Among all the KAN variants which have been developed, FastKAN has attracted particular attention because it outperforms in terms of efficiency and accuracy. It significantly reduces the computational cost by replacing the spline bases  with Gaussian RBFs.
However, a critical component of RBF-based models is the \emph{shape parameter}, which controls the width of the RBF.  The choice of this parameter strongly influences approximation accuracy: small values may lead to overfitting, whereas large values can oversmooth the solution and degrade performance. 
In the standard FastKAN formulation, the shape parameter is fixed, which limits the adaptability of the model across different types of target functions.

In this work, we propose an adaptive extension of FastKAN that addresses the limitations associated with fixed kernel choices and fixed shape parameters. 
Specifically, we introduce an \emph{adaptive RBF-KAN} framework --- also referred to as \emph{adaptive FastKAN} to emphasize its computational efficiency --- in which the initial kernel scale is estimated using leave-one-out cross-validation (LOOCV) and subsequently refined during network training. 
The LOOCV procedure provides a data-driven initialization for the RBFs underlying the KAN edge representations, while gradient-based optimization allows the parameter to adapt to the structure of the deep network. 
To the best of our knowledge, this is the first work that connects LOOCV-based kernel scale estimation with the training of deep KAN architectures.

In addition, we extend the FastKAN formulation by incorporating a broader family of radial basis kernels beyond the Gaussian basis typically used in FastKAN. 
In particular, we introduce Matérn and Wendland kernels into the KAN framework for the first time. 
These kernels offer different smoothness and locality properties, allowing the network to better adapt its basis representation to functions with varying structural characteristics.

To evaluate the proposed approach, we conduct numerical experiments on four benchmark functions and compare the results with several baseline architectures, including MLPs, the standard KAN, and existing FastKAN variants. 
The benchmark functions represent different approximation challenges, including smooth surfaces, discontinuities, oscillatory behavior, and localized singularities. 
The experiments illustrate that both kernel selection and LOOCV-based initialization play an important role in the performance of RBF-based KAN models, with different kernels proving more suitable for different classes of functions. 

The remainder of the paper is organized as follows. Section~\ref{sec:background} reviews the KA representation theorem and relevant background. Section~\ref{sec:kan_versions} summarizes several existing KAN architectures. 
Section~\ref{sec:methodology} introduces the proposed adaptive RBF-KAN method and describes the training procedure. 
Section~\ref{sec:results} presents numerical experiments and comparisons with other models. 
Finally, Section~\ref{sec:conclusion} concludes the paper and discusses possible directions for future work.

\section{Background and Related Work} \label{sec:background}
\subsection{Kolmogorov-Arnold representation theorem}
There has been much discussion on the benefits of having more hidden layers in a neural network.  The KA representation theorem \cite{kolmogorov56} offers a possible solution by showing that every continuous function may be represented by a network with two hidden layers \cite{hecht1987}.  In the literature, there has been much discussion over this view.  Some articles, for example \cite{girosi1989}, contend that the KA theorem and neural networks are not as related as they are claimed to be, while others, as \cite{kurkova1991}, support its relevance to neural network design.

The original version of the KA representation theorem states that for any continuous function \( f : [0, 1]^d \to \mathbb{R} \), there exist univariate continuous functions $\eta_n : \mathbb{R} \to \mathbb{R}$ and $\psi_{m,n} : [0, 1] \to \mathbb{R}$ such that:
\begin{align}\label{KA_THM}
f(x_1, \dots, x_d) &= \sum_{n=1}^{2d+1} \eta_n \left( \sum_{m=1}^{d} \psi_{m,n}(x_m) \right).
\end{align}

This formulation shows that \( (2d + 1)  \) univariate functions are sufficient to exactly represent a \( d \)-variate function. Kolmogorov published this result in 1956~\cite{kolmogorov56}, effectively disproving Hilbert's 13th problem, which was concerned with the solution of algebraic equations. The introduction of multiple layers in neural networks, motivated in part by the KA representation theorem, dates back to the 1960s, although the connection between the two concepts emerged much later \cite{schmidt2021}.

\subsection{A mathematical overview of KAN}
KANs are neural network models that are constructed using the KA representation theorem \cite{kolmogorov56}.   The innovative concept behind KAN is to replace the conventional fixed linear weights used in neural networks with learnable univariate functions, which are frequently expressed using B-splines.   This technique allows KANs to capture complex nonlinear behavior of several functions, leading to increased flexibility and accuracy.

A series of compositions of univariate functions applied to linear combinations of inputs may be used to mathematically characterize the basic building blocks of KANs. The expression for the forward propagation of a KAN with \( L \) layers is 
\begin{align}\label{kan-simple}
x_{m}^{(k+1)} &= \sum_{n=1}^{d_k} \psi_{k,m,n}(x_n^{(k)}),
\end{align}
where \( \psi_{k,m,n} \) represents the learnable univariate function connecting neuron \( n \) in layer \( k \) to neuron \( m \) in layer \( k + 1 \). The parameterization of the univariate functions in the typical KAN implementation relies on B-spline basis functions. Specifically, each function \( \psi_{k,m,n}(x) \) is given by:
\begin{align}\label{Silu-psi}
\psi_{k,m,n}(x) &= \sum_{j=0}^{H+J-1} c_{k,m,n,j} B_{j,J}(x) + w_{k,m,n} \cdot \text{SiLU}(x).
\end{align}
Here, \( B_{j,J}(x) \) represents the B-spline basis functions of degree \( J \), \( c_{k,m,n,j} \) are the learnable spline coefficients, and \( H \) denotes the number of grid intervals. As the SiLU activation function \cite{liu24} is applied, we can observe its residual connection with the second term, \( w_{k,m,n} \cdot \text{SiLU}(x) \).

The practical implementation of KAN is truly aligned with the original KA theorem. Although the theorem specifies that exactly \( 2d + 1 \) outer functions are required for a \( d \)-dimensional input, contemporary KAN architectures frequently disregard this condition, employing arbitrary network widths and depths \cite{yu24}. In practice, this means KANs are not as different from MLPs as they are often claimed to be. Dropping the strict $2d+1$ constraint and using arbitrary widths and depths makes them look a lot like MLPs with learnable activation functions rather than fixed ones. Moreover, the mathematical equivalence between KANs and certain classes of RBFNs and tensor product spline methods has been largely overlooked in the literature \cite{mont20}. This equivalence implies that KANs might be better viewed as a reinterpretation of established techniques rather than a groundbreaking innovation, prompting a reassessment of the novelty claims that have fueled much of the recent enthusiasm.

\section{Versions of Kolmogorov-Arnold Networks} \label{sec:kan_versions}
In this section we will provide an overview of various versions of KANs by giving the details about their formation and also revealing the facts that these versions perform better than the basic KANs in some specific problems.

\subsection{FastKAN}
As discussed earlier, KAN is a network that works on the principle of using learnable univariate nonlinear activation functions which are usually B-splines. The latter make the network computationally expensive. To address this issue, Li \cite{li24} introduced a new version of KAN, known as FastKAN. The main idea behind this version is to approximate the 3-order B-spline basis using the Gaussian RBF kernel. In addition to this, to avoid deviation of the input from the RBF domain, layer normalization is used \cite{ba16}. By utilizing these simple modifications in the implementation of KAN, experiments have shown that this version make it much faster than the original one so it has been named as FastKAN.

A set of real valued-functions that depend only on the values determined by the radial distance, or the distance from a central point, are referred as RBFs \cite{buh2000}. They have many diverse applications including machine learning, function approximation, and pattern recognition. The working principle of RBFs is to join many radially symmetric functions that are centered at various locations in the input space to approximate a function. A linear combination of these RBFs, weighted by adjustable coefficients, is the output of an RBFN. The mathematical representation of a RBF network with $K$ centers is given by
\begin{align}
\label{rbf}
f(x) &= \sum_{k=1}^{K} w_k\phi(|x - c_k|),
\end{align}
where $x$ is the input value, $c_k$ are the centers, $|x - c_k|$ is Euclidean distance between them and $w_k$ are the adaptive weights. As discussed earlier, FastKAN uses the Gaussian RBF as a basis function,
\begin{align} \label{ga_rbf}
\phi(r) = \exp\left( -\frac{r^2}{2h^2} \right)
\end{align}
where $h$ is a parameter that regulates the spread of the function and $r = |\cdot|$ is the Euclidean distance defined in \eqref{rbf}. Similarly to Gaussian, we have also tested FastKAN for popular strictly positive definite RBFs \cite{fasshauer15,wen05} with various order of smoothness (see Table \ref{tab_rbf}).

\begin{table}[h]
\centering
\begin{tabular}{ll}
\hline
\rule[0mm]{0mm}{3ex}
RBF  & $\phi(r)$ \\

\hline
\vspace{2mm}
\rule[0mm]{0mm}{3ex}
{Gaussian $C^{\infty}$} (GA) & $\exp\left( -\frac{r^2}{2h^2} \right)$   \\
\vspace{2mm}
\rule[0mm]{0mm}{3ex}
{Inverse MultiQuadric $C^{\infty}$} (IMQ) & $\left( 1 + \frac{r^2}{h^2} \right)^{-1/2}$   \\
\vspace{2mm}
\rule[0mm]{0mm}{3ex}
{Mat$\acute{\text{e}}$rn $C^6$} (M6)  & $\exp\left( -\frac{r}{h} \right) \left( \frac{r^3}{h^3} + 6\frac{r^2}{h^2} + 15\frac{r}{h} + 15 \right)$   \\
\vspace{2mm}
\rule[0mm]{0mm}{3ex}
{Mat$\acute{\text{e}}$rn $C^4$} (M4)  & $\exp\left( -\frac{r}{h} \right) \left( \frac{r^2}{h^2} + 3\frac{r}{h} + 3 \right)$   \\
\vspace{2mm}
\rule[0mm]{0mm}{3ex}
{Mat$\acute{\text{e}}$rn $C^2$} (M2) & $\exp\left( -\frac{r}{h} \right) \left( \frac{r}{h} + 1 \right)$   \\
\vspace{2mm}
\rule[0mm]{0mm}{3ex}
{Wendland $C^6$} (W6)  & $\max\left( 1 - \frac{r}{h}, 0 \right)^8 \left( 32\frac{r^3}{h^3} + 25\frac{r^2}{h^2} + 8\frac{r}{h} + 1 \right)$  \\
\vspace{2mm}
\rule[0mm]{0mm}{3ex}
{Wendland $C^4$} (W4)  & $\max\left( 1 - \frac{r}{h}, 0 \right)^6 \left( 35\frac{r^2}{h^2} + 18\frac{r}{h} + 3 \right)$  \\
\vspace{2mm}
\rule[0mm]{0mm}{3ex}
{Wendland $C^2$} (W2) & $\max\left( 1 - \frac{r}{h}, 0 \right)^4 \left( 4\frac{r}{h} + 1 \right)$  \\

\hline
\end{tabular}
\caption{Examples of RBFs used in FastKAN.}
\label{tab_rbf}
\end{table}

\subsection{EfficientKAN}
EfficientKAN \cite{blealtan24} is an optimized implementation of KAN designed to address the performance issues of the original KAN implementation. The key performance issue in the original implementation arises from the need to expand intermediate variables for activation functions, which increases memory and computational cost. EfficientKAN resolves this by reformulating the computation as matrix multiplication, resulting in significant improvements in speed and memory efficiency.

The key innovations of EfficientKAN include:
\begin{itemize}
    \item \emph{Reformulation of Computation}: It replaces the tensor expansion in the original KAN with a simpler matrix multiplication approach by applying the input to different basis functions (B-splines) and then combining them linearly.
    \item \emph{Memory Efficiency}: This reformulation drastically lowers memory expenses, improving the computational efficiency of the model.
     \item \emph{Regularization}: Unlike the original KAN that employed $L_1$ regularization on input samples, this approach employs $L_1$ regularization on the weights, which is consistent with the reformulated algorithm.
     \item \emph{Learnable Activation Functions and Scale}: Each activation function has the option of learnable scales; if desired, this can be turned off for efficiency.
\end{itemize}

\subsection{Kolmogorov-Arnold-Fourier network}
The Kolmogorov-Arnold-Fourier (KAF) network \cite{Zhang2025} builds on KAN by addressing two practical issues: the parameter overhead that grows with input dimension, and the difficulty of capturing high-frequency features in complex tasks. 

To balance parameter efficiency with spectral representation, KAF combines two components: trainable Random Fourier Features (RFF) and a hybrid GELU-Fourier activation function. Here, GELU (Gaussian Error Linear Unit) is defined as $\text{GELU}(x) = x \cdot \Phi(x)$, where $\Phi$ is the cumulative distribution function of the standard normal distribution~\cite{hendrycks2016}. It has proven effective in various domains, including differential equation solving, computer vision, natural language processing, and audio processing.

\begin{figure}[h!]
    \centering
    \includegraphics[width=0.6\textwidth]{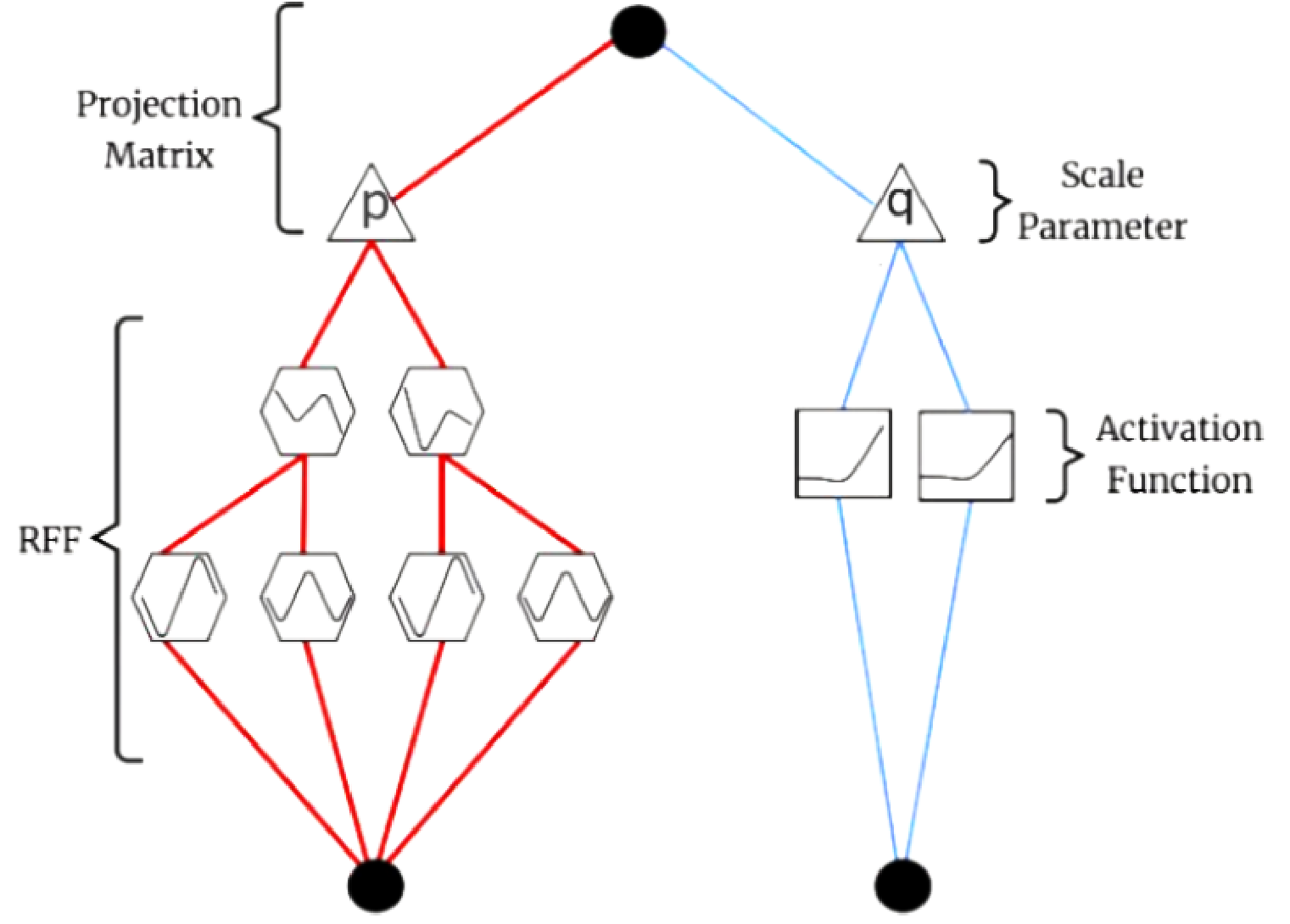}
   \caption{KAF layer architecture.}
    \label{fig:KAF}
\end{figure}

Some of the main innovative features of KAF are:
\begin{itemize}
    \item \emph{Dual-Matrix Structure}: By utilizing the properties of matrix association, KAF significantly reduces the number of parameters compared to KAN.
    
    \item \emph{Learnable RFF Initialization}: The spectral distortion is eliminated by learnable RFF initilization, which is especially advantageous for high-dimensional approximation applications.
    \item \emph{Adaptive Hybrid Activation}: GELU-Fourier activation progressively improves frequency representation during training.
\end{itemize} 
For compactness, and consistent with the original KAF formulation, we use vectorized notation in this subsection, where bold symbols denote vectors and matrices.
In particular, the main computational procedure for each KAF layer can be expressed as follows:
\begin{equation}
\mathbf{O}^{(k)} = 
\underbrace{\mathbf{U}^{(k)}}_{\text{output projection}}
\left(
    \underbrace{
        \mathbf{p}^{(k)} \odot \text{GELU}(\mathbf{z}^{(k)}) + \mathbf{q}^{(k)} \odot \psi(\mathbf{z}^{(k)})
    }_{\text{modulated feature composition}}
\right)
+ \mathbf{d}^{(k)},
\label{eq:afl_layer}
\end{equation}
where $\mathbf{z}^{(k)} = \text{LayerNorm}(\mathbf{x}^{(k)})$ denotes the layer-normalized input at layer $k$~\cite{ba16}, $\psi(\cdot)$ is the nonlinear mapping based on RFF, $\mathbf{p}^{(k)}, \mathbf{q}^{(k)} \in \mathbb{R}^n$ are learnable scaling parameters that scale the contributions of the RFF features and GELU activation, respectively, while $\mathbf{U}^{(k)} \in \mathbb{R}^{m \times n} $ is the linear transformation weight and $\mathbf{d}^{(k)} \in \mathbb{R}^m$ is the output bias vector. The overall structure of a KAF layer is illustrated in Figure~\ref{fig:KAF}.

KAF has been shown to outperform existing methods in various domains, combining interpretability with computational efficiency.

\subsection{Chebyshev polynomial-based KAN}
To increase the effectiveness and precision of the approximation for nonlinear functions a novel architecture, called the Chebyshev Kolmogorov-Arnold Network (Cheby-KAN) \cite{sidharth2024}, was developed. The Cheby-KAN provides notable benefits over conventional MLPs. It is motivated by the KA theorem and the potent approximation capabilities of Chebyshev polynomials. 
 The central Chebyshev polynomials, a class of orthogonal polynomials defined on the interval [$-1, 1$], are an excellent choice for function approximation within this framework.  Based on these polynomials, the Cheby-KAN layer provides a unique substitute for the original B-splines, overcoming their drawbacks in terms of performance and usability. The replaced splines with the global orthogonal polynomial in Cheby-KAN can be expressed mathematically as:
\begin{equation*}
\psi_{k,m,n}(x) = \sum_{j=0}^{P} c_{j} C_j(x), 
\end{equation*}
where
\begin{equation*}
C_j(x) = \cos(j \arccos(x)), \quad x \in [-1, 1].
\end{equation*}
Here, $C_j(x)$ represents a Chebyshev polynomial of the first kind that helps in minimization of maximum error in the $L_\infty$ norm, resulting in an exponential convergence near-optimal polynomial approximation for smooth functions \cite{lin24,gao24}.

\begin{figure}[h!]
    \centering
    \includegraphics[width=0.6\textwidth]{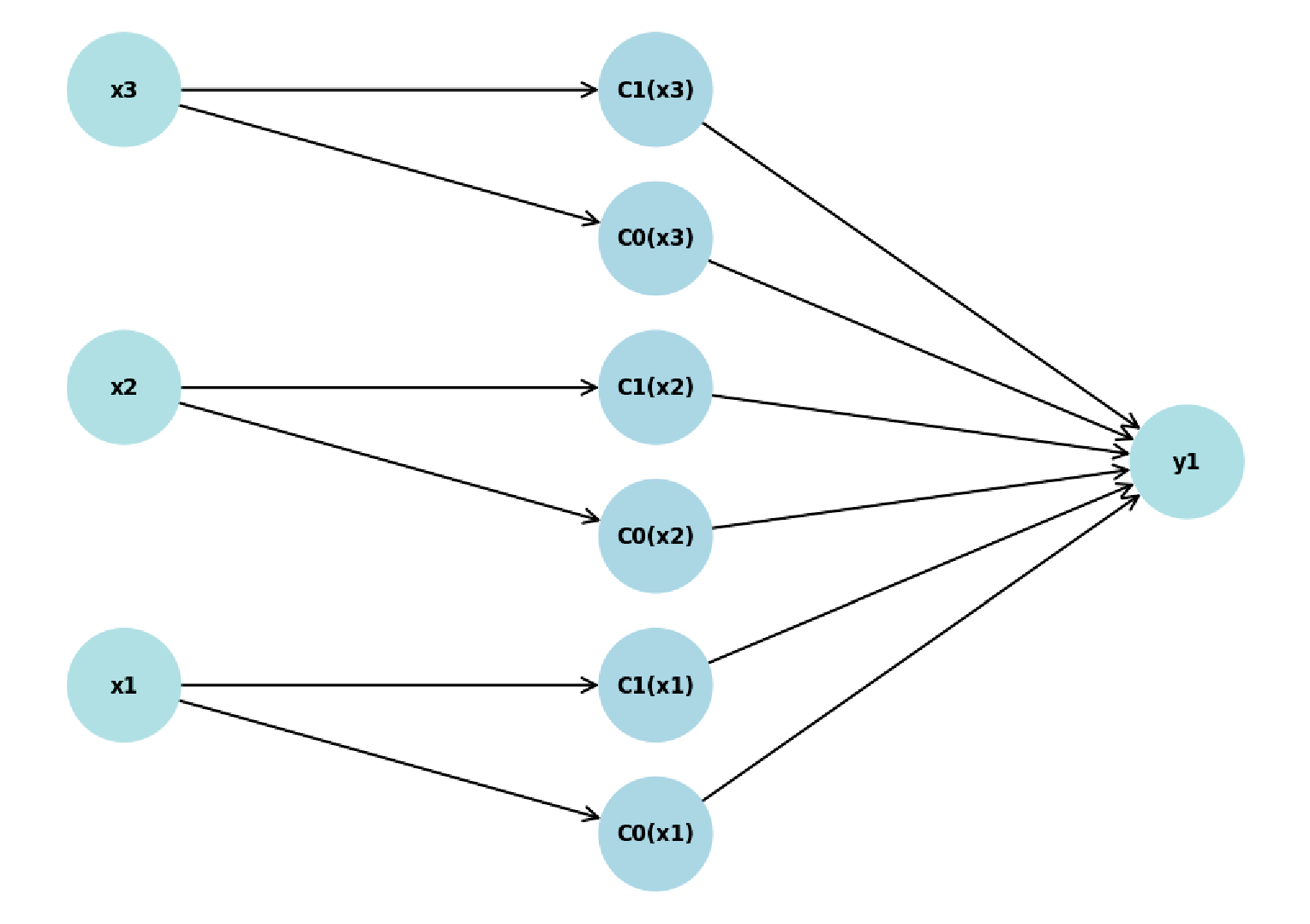}
    \caption{Cheby-KAN architecture.}
    \label{fig:cheby-kan}
\end{figure}

The Chebyshev KAN model with three input features, the Chebyshev polynomial of degree 1, and output shape 1 is visualized in Figure \ref{fig:cheby-kan}. However, the image does not display the weights or coefficients. 
\begin{table}[H]
    \centering
    \caption{Comparison between basic KAN and Cheby-KAN.}
    \label{tab:kan_comparison}
    \begin{tabular}{l p{5.5cm} p{5.5cm}}
        \toprule
        \textbf{Feature} & \textbf{Basic KAN (spline-based)} & \textbf{Cheby-KAN} \\
        \midrule
        Domain & Local (small knot-defined intervals) & Global, confined to $x \in [-1, 1]$ \\
        \addlinespace 
        Flexibility & High adaptibilty for local fluctuations & Restricted, smooth global approximation \\
        \addlinespace
        Stability/Robustness & Stable (insensitive to refinement of knots) & Stable only inside $[-1, 1]$ \\
        \addlinespace
        Parameters Setting & Need coefficients $c_{k,m,n,j}$ and knot placement & Need just polynomial coefficients $c_j$ \\
        \addlinespace
        Ideal for & Scenarios having sharp local variations & Functions that are smooth on bounded domains \\
        \bottomrule
    \end{tabular}
\end{table}

\section{Methodology} \label{sec:methodology}
As mentioned above, the KA representation theorem \cite{kolmogorov56} served as  inspiration for this innovative neural network design methodology of KANs.  The primary characteristic of KAN is its ability to substitute learnable univariate functions for conventional fixed linear weights.  This development enables KANs to effectively model complicated nonlinear functions, improving interpretability and accuracy. By parameterizing the univariate functions \( \psi \) as B-spline curves with learnable coefficients, Liu et al. \cite{liu24} expanded \eqref{KA_THM}. However, only a two-layer KAN is described by \eqref{KA_THM}, which might not be descriptive enough for complicated problems. This restriction can be reduced by implementing deeper structures that resemble MLPs.

For example, say that we are dealing with a supervised learning project with input-output pairs $\{\mathbf{x}_k, y_k\}$, and we want to determine $f$ such that for every data point, $y_k \approx f(\mathbf{x}_k)$. If suitable univariate functions $\psi_{m,n}$ and $\eta_n$ can be identified, then \eqref{KA_THM} provides an exact representation of $f$, and the approximation problem is formally resolved. This motivated Liu et al. \cite{liu24} to create a neural network that explicitly parametrizes \eqref{KA_THM}.  They stated that they can parametrize each 1D function as a B-spline curve with learnable coefficients of local B-spline basis functions because all the functions to be learned are univariate functions. 

In order to construct deep KANs, we must first address the question of what is a KAN layer.  It turns out that the following matrix of 1D functions may be used to build a KAN layer with $d_{\text{in}}$-dimensional inputs and $d_{\text{out}}$-dimensional outputs:
\begin{equation}\label{psi-fun}
    \boldsymbol{\Psi} = \{ \psi_{m,n} \}, \quad n = 1, 2, \ldots, d_{\text{in}}, \quad m = 1, 2, \ldots, d_{\text{out}}, 
\end{equation}
where the parameters of the functions $\psi_{m,n}$ are trainable.  The inner functions in the KA theorem form a KAN layer with $d_{\text{in}} = d$ and $d_{\text{out}} = 2d + 1$, whereas the outer functions form a KAN layer with $d_{\text{in}} = 2d + 1$ and $d_{\text{out}} = 1$. Therefore, the KA representations in \eqref{KA_THM} are just two KAN layer composites.  The definition of deeper KA representations is now evident and it tells us to just add more KAN layers. A standard KAN network consists of $L$ layers. Specifically, let $\boldsymbol{\Psi}_k$ denote the $k$-th KAN layer defined as in \eqref{psi-fun}, with $d_{\mathrm{in}} = d_k$ and 
$d_{\mathrm{out}} = d_{k+1}$, for $k = 0, 1, \ldots, L-1$. Given an input vector $\mathbf{x} \in \mathbb{R}^{d_0}$, the KAN output is:
\begin{equation}
  \text{KAN}(\mathbf{x}) = (\boldsymbol{\Psi}_{L-1} \circ 
  \boldsymbol{\Psi}_{L-2} \circ \cdots \circ \boldsymbol{\Psi}_1 \circ 
  \boldsymbol{\Psi}_0)\mathbf{x}. 
  \label{Kan-out}
\end{equation}
On the other hand, let $\boldsymbol{\Theta}_k : \mathbb{R}^{d_k} \to \mathbb{R}^{d_{k+1}}$ denote indexed affine maps for $k = 0, 1, \ldots, L-1$. An MLP can then be written as a composition of these maps with fixed non-linearities $\gamma$:
\begin{equation}
    \text{MLP}(\mathbf{x}) = (\boldsymbol{\Theta}_{L-1} \circ \gamma 
    \circ \boldsymbol{\Theta}_{L-2} \circ \gamma \circ \cdots \circ 
    \boldsymbol{\Theta}_1 \circ \gamma \circ 
    \boldsymbol{\Theta}_0)\mathbf{x}.
    \label{MLP}
\end{equation}
Comparing \eqref{Kan-out} and \eqref{MLP}, the key difference is that each $\boldsymbol{\Psi}_k$ handles both linearity and nonlinearity 
jointly, while MLP separates them into the affine maps $\boldsymbol{\Theta}_k$ and the fixed activations $\gamma$. As a result, backpropagation can be used to train KANs. The latter are merely combinations of splines and MLPs, which efficiently leverage on each other's advantages while limiting their respective disadvantages, despite their complicated mathematical basis. Splines are notable for their ability to transition between different resolutions and their accuracy on low-dimensional functions. 

KANs with a finite grid size can overcome the curse of dimensionality by accurately approximating the function with a residue rate that is independent of the dimension. Since splines are exclusively used to approximate 1D functions, this makes sense.

The implementation steps of KAN architecture that we have used for the approximation of the 2D function are briefly discussed in Algorithm \ref{alg1}. 

\begin{algorithm} 
\caption{KAN for Function Approximation} \label{alg1}
\begin{algorithmic}

\State \textbf{Input:} Training set $\mathcal{D}_{train} = \{(\mathbf{x}_i, y_i)\}_{i=1}^{N_{train}}$, test set $\mathcal{D}_{test} = \{(\mathbf{x}_i, y_i)\}_{i=1}^{N_{test}}$, layer widths sequence $\mathbf{d} = [d_0, d_1, \ldots, d_L]$, grid intervals $H$, spline degree $J$, learning rate $\alpha$, epochs $E$, where each layer $\boldsymbol{\Psi}_k = \{\psi_{k,m,n}\}$ is defined as in \eqref{psi-fun}

\State \textbf{Output:} Trained model 
$\{\boldsymbol{\Psi}_k\}_{k=0}^{L-1}$, 
predicted values $\hat{\mathbf{y}}$ over 
the domain

\Statex
\State \textbf{Initialization:}
\State \quad Initialize learnable spline 
coefficients $c_{k,m,n,j}$ and residual weights 
$w_{k,m,n}$ as defined in \eqref{Silu-psi}, 
for all $\psi_{k,m,n} \in \boldsymbol{\Psi}_k$, 
$k = 0,\ldots,L-1$

\Statex
\State \textbf{Training:}
\For{epoch $e = 1$ to $E$}
    \State \textbf{Forward pass 
    (see \eqref{Kan-out}):}
    \State \quad Set $\mathbf{x}^{(0)} = \mathbf{x}$
\Comment{superscript denotes layer index}
    \For{layer $k = 0$ to $L-1$}
        \State \quad For each 
        $m \in \{1,\ldots,d_{k+1}\}$:
        \State \quad $x^{(k+1)}_m = 
        \sum_{n=1}^{d_k} 
        \psi_{k,m,n}(x^{(k)}_n)$
        \Comment{see \eqref{kan-simple}}
        \State \quad where each edge function is:
        \State \quad $\psi_{k,m,n}(x) = 
        \sum_{j=0}^{H+J-1} 
        c_{k,m,n,j} B_{j,J}(x)$
    \EndFor
    \State \quad $\hat{y}_i = x^{(L)}$
    \Comment{predicted value at current epoch}

    \Statex
    \State \textbf{Backpropagation:}
    \State \quad $\mathcal{L} = 
    \frac{1}{N_{train}} 
    \sum_{i=1}^{N_{train}} 
    (y_i - \hat{y}_i)^2$
    \Comment{MSE loss}
    \State \quad $\{c_{k,m,n,j}, w_{k,m,n}\} 
    \leftarrow \{c_{k,m,n,j}, w_{k,m,n}\} - 
    \alpha \cdot \nabla \mathcal{L}$
    \Comment{Adam update}
    \If{$e \bmod 100 = 0$}
        \State \quad Evaluate on 
        $\mathcal{D}_{test}$, record loss
    \EndIf
\EndFor

\Statex
\State \textbf{Surface Reconstruction:}
\State \quad Evaluate $\hat{y}_i = 
\text{KAN}(\mathbf{x}_i)$ via \eqref{Kan-out} 
for each point $\mathbf{x}_i$ in the domain 
to obtain $\hat{\mathbf{y}}$
\Comment{grid details in 
Section~\ref{sec:results}}

\State \Return 
$\{\boldsymbol{\Psi}_k\}_{k=0}^{L-1}$, 
$\hat{\mathbf{y}}$

\end{algorithmic}
\end{algorithm}

\subsection{Adaptive RBF-KAN} 

FastKAN replaces the spline basis functions of the original KAN architecture with RBFs, typically Gaussian kernels \cite{li24}. 
As a result, each edge function $\psi_{k,m,n}$ in \eqref{kan-simple} can be written as a one-dimensional kernel expansion over fixed grid centers. 
Specifically, the functional representation on each edge takes the form
\begin{equation}
\psi_{k,m,n}(x)
=
\sum_{j=1}^{K}
c_{k,m,n,j}
\,
\phi\!\left(\frac{|x-c_j|}{h}\right),
\end{equation}
where $\{c_j\}_{j=1}^{K}$ are grid centers, $c_{k,m,n,j}$ are learnable coefficients, and $\phi$ is the radial kernel defined in \eqref{ga_rbf}.

This representation is structurally identical to classical kernel approximation models, where the function is expressed as a linear combination of localized basis functions centered at fixed nodes \cite{fasshauer07,Amir24}. 
Consequently, the coefficients $c_{k,m,n,j}$ play the same role as kernel weights in standard RBF interpolation \cite{cavoretto2021,cavoretto2024}, at least at the level of the one-dimensional edge expansion.

Importantly, although the complete KAN model is a deep composition of such edge functions, the functional building block remains a \emph{one-dimensional kernel expansion}. 
Therefore, classical kernel parameter selection strategies can still provide useful information about the appropriate scale of the basis functions. 
In particular, the kernel shape parameter $h$ determines the spatial resolution of the basis functions and directly affects the approximation properties of each edge function.

To obtain a principled initialization for this parameter, we estimate the kernel scale using a LOOCV criterion applied to an auxiliary one-dimensional kernel expansion model corresponding to the functional representation above. 
In this setting, the coefficients play the role of interpolation weights for the auxiliary RBF model, while the kernel parameter $h$ determines the effective bandwidth of the basis functions.

\subsubsection{Finding the Initial Shape Parameter via LOOCV}

Given a set of $N$ one-dimensional points $\{x_i, y_i\}_{i=1}^{N}$ 
used to construct the auxiliary kernel interpolation system for shape parameter 
estimation, classical RBF interpolation builds an approximation of 
the form
\begin{equation}
s(x) = \sum_{j=1}^{N} w_j\phi\!\left(\frac{|x-x_j|}{h}\right),
\end{equation}
where $\phi(\cdot)$ is the chosen radial kernel as defined in Table~\ref{tab_rbf}, and $\mathbf{w} = (w_1,\ldots,w_N)^\top$ are 
interpolation coefficients.

In our implementation, the one-dimensional points $\{x_i\}_{i=1}^N$ are taken from a single input coordinate, while $\{y_i\}_{i=1}^N$ denote the corresponding target values used in this auxiliary initialization step. Since all input coordinates are scaled in the same way, we observed that different coordinate choices lead to very similar LOOCV minima, so a single representative coordinate is sufficient for estimating an initial kernel scale.

The interpolation coefficients are obtained by solving the linear system
\begin{equation}
A \mathbf{w} = \mathbf{y},
\end{equation}
where $\mathbf{y} = (y_1,\ldots,y_N)^\top \in \mathbb{R}^N$, 
$\mathbf{w} = (w_1,\ldots,w_N)^\top \in \mathbb{R}^N$, and 
$A \in \mathbb{R}^{N \times N}$ is the interpolation matrix with entries
\begin{equation}
A_{ij} = \phi\!\left(\frac{|x_i-x_j|}{h}\right).
\end{equation}

A common approach for selecting the kernel scale is LOOCV. A direct use of LOOCV would require solving $N$ interpolation systems, each leaving out one data point. However, Rippa \cite{rippa99} showed that, for the unregularized interpolation problem, the LOOCV errors can be computed efficiently using a single matrix inversion. In this case, if $A$ is invertible, the leave-one-out prediction 
error for the $i$-th data point can be 
computed directly as
\begin{equation}
e_i = \frac{w_i}{(A^{-1})_{ii}},
\end{equation}
where $\mathbf{w} = A^{-1}\mathbf{y}$ are 
the interpolation coefficients.
This result allows the LOOCV error to be evaluated for all data points simultaneously. In our approach we evaluate this LOOCV error for a range of candidate values of the kernel parameter $h$. 

For numerical stability, the interpolation matrix $A$ is replaced by $A + \lambda I$, where $I \in \mathbb{R}^{N \times N}$ is the identity matrix and $\lambda = 10^{-9}$ is a small perturbation chosen to prevent ill-conditioning of $A$ when data points are close together or the kernel is nearly flat. The same 
regularization is applied in Algorithm~\ref{alg2}. In this regularized setting, we use the same expression as a stabilized LOOCV-type score for selecting the initial value of $h$. Thus, the resulting procedure should be interpreted as a numerically robust initialization criterion rather than as an exact optimization principle for the full deep KAN.

The value of $h$ minimizing the LOOCV criterion is 
denoted $h_{opt}$ and is returned by 
Algorithm~\ref{alg2}. This value is then used 
to initialize the network shape parameter, 
setting $h_{init} = h_{opt}$ at the start of 
training in Algorithm~\ref{alg3}.
It is important to emphasize that this optimization is performed only to estimate the characteristic scale of the kernel basis functions. 
Because the final predictor is a deep KAN architecture formed by compositions of such one-dimensional expansions, the LOOCV estimate is not expected to be globally optimal for the entire network. 
Instead, it provides a data-driven initialization that reflects the appropriate spatial resolution of the kernel basis.

During network training the kernel parameter is treated as a learnable variable and updated through backpropagation together with the coefficients $c_{k,m,n,j}$. 
Empirically, we observe that the trained value typically remains close to the LOOCV estimate, indicating that the initialization captures a suitable kernel scale while significantly reducing the need for expensive random hyperparameter search. The two-stage LOOCV search for $h_{opt}$ is summarized in Algorithm~\ref{alg2}, and the complete adaptive RBF-KAN training procedure is described in Algorithm~\ref{alg3}.

\subsubsection{Shape parameter optimization}

LOOCV gives us a very good static starting point. However, when the FastKAN is training, the RBF coefficients and other edge parameters change, so the ideal RBF width might also need to be adaptive according to the nature of the problem. We want to continuously optimize the shape parameter using gradient descent. 

The main problem is that $h$ must remain strictly positive. If we allow the optimizer to update $h$ directly, a large gradient step could drive $h$ to zero or to a negative value. 

A value of $h = 0$ leads to division by zero in the kernel evaluation, while a negative value $h < 0$ causes distance-based kernels such as Mat\'{e}rn and Wendland to grow exponentially 
rather than decay, resulting in severe numerical instability. To prevent this, we enforce positivity by mapping the shape parameter into log space with a new learnable parameter $\theta = \ln(h)$.

Before starting training, we initialize 
$\theta = \ln(h_{opt})$, where $h_{opt}$ 
is the value returned by Algorithm~\ref{alg2}. Then, during every forward pass in the training epochs, the network calculates the real shape back using the exponential function, i.e., $h = e^{\theta}$.

Because $e^{\theta}$ is always positive for any real number $\theta$, the Adam optimizer can update $\theta$ in any direction using backpropagation. There is no risk of violating the RBF constraints. By treating $\theta$ just like the other learnable RBF coefficients on the network edges, the model minimizes the loss and dynamically adapts the RBF shape to the data at the same time.

\begin{algorithm}
\caption{Two-Stage LOOCV Search for Initial 
Shape Parameter} \label{alg2}
\begin{algorithmic}

\State \textbf{Input:} One-dimensional data 
$\{x_i, y_i\}_{i=1}^{N}$, kernel $\phi$ 
(see Table~\ref{tab_rbf}), search bounds 
$h_{\min}$, $h_{\max}$, coarse points 
$N_{coarse}$, fine points $N_{fine}$, 
regularization parameter $\lambda$

\State \textbf{Output:} Optimal initial shape 
parameter $h_{opt}$

\Statex
\State Compute pairwise distances 
$D_{ij} \gets |x_i - x_j|$
\Comment{Absolute value: 1D scalar inputs}

\State $h_{opt} \gets h_{\min}$
\Comment{Optimal shape parameter}
\State $err_{min} \gets \infty$
\Comment{Minimum LOOCV error found so far}

\Statex
\State \textbf{Stage 1 --- Coarse search:}
\For{each $h$ in $N_{coarse}$ equally spaced 
points in $[h_{\min}, h_{\max}]$}
    \State $A_{ij} \gets \phi(D_{ij}/h)$
    \Comment{Interpolation matrix, 
    Table~\ref{tab_rbf}}
    \State $A \gets A + \lambda I$
    \Comment{$\lambda$ defined in 
    Section~\ref{sec:methodology}}
    \State $\mathbf{w} \gets A^{-1}\mathbf{y}$
    \Comment{Interpolation coefficients}
    \State $e_i \gets w_i/(A^{-1})_{ii}$ 
    for all $i$
    \Comment{Rippa's formula}
    \State $err \gets \max_i |e_i|$
    \If{$err < err_{min}$}
        \State $h_{opt} \gets h$
        \State $err_{min} \gets err$
    \EndIf
\EndFor

\Statex
\State \textbf{Stage 2 --- Fine search 
around $h_{opt}$:}
\State $r_{h} \gets 
2 \times (h_{\max} - h_{\min})/N_{coarse}$
\Comment{Fine search half-width: 2 coarse steps}
\For{each $h$ in $N_{fine}$ equally spaced 
points around $h_{opt}$ within $r_{h}$}
    \State $A_{ij} \gets \phi(D_{ij}/h)$,
    \quad $A \gets A + \lambda I$
    \State $\mathbf{w} \gets A^{-1}\mathbf{y}$
    \State $e_i \gets w_i/(A^{-1})_{ii}$ 
    for all $i$
    \State $err \gets \max_i |e_i|$
    \If{$err < err_{min}$}
        \State $h_{opt} \gets h$
        \State $err_{min} \gets err$
    \EndIf
\EndFor

\State \Return $h_{opt}$

\end{algorithmic}
\end{algorithm}

\begin{algorithm}
\caption{Adaptive RBF-KAN Training} \label{alg3}
\begin{algorithmic}

\State \textbf{Input:} Training set 
$\mathcal{D}_{train} = \{(\mathbf{x}_i, y_i)\}_{i=1}^{N_{train}}$, 
test set 
$\mathcal{D}_{test} = \{(\mathbf{x}_i, y_i)\}_{i=1}^{N_{test}}$, 
kernel $\phi$ (see Table~\ref{tab_rbf}), 
layer widths $\mathbf{d} = [d_0, d_1, \ldots, d_L]$, 
number of RBF grid points $K$ with centers in 
$[c_{\min}, c_{\max}]$, 
learning rate $\alpha$, 
total epochs $E$

\State \textbf{Output:} Trained model 
$\{\boldsymbol{\Psi}_k\}_{k=0}^{L-1}$ 
and predicted values $\hat{\mathbf{y}}$ 
over the domain

\Statex
\State \textbf{Step 1: Initialize shape parameter}
\State Compute $h_{opt}$ via Algorithm~\ref{alg2}
\Comment{Data-driven initialization using LOOCV}
\State $\theta \gets \ln(h_{opt})$
\Comment{Log-space reparametrization ensures $h = e^\theta > 0$ throughout training}

\Statex
\State \textbf{Step 2: Initialize network}
\State Set grid centers $\{c_j\}_{j=1}^{K}$ 
equally spaced in $[c_{\min}, c_{\max}]$
\State Initialize coefficients $c_{k,m,n,j}$ and 
residual weights $w_{k,m,n}$ for all 
$\psi_{k,m,n} \in \boldsymbol{\Psi}_k$, 
$k = 0, \ldots, L-1$
\Comment{See \eqref{psi-fun}}
\State Configure Adam optimizer with learning 
rate $\alpha$ to update 
$\theta$ and $\{c_{k,m,n,j}, w_{k,m,n}\}$

\Statex
\State \textbf{Step 3: Train}
\For{epoch $e = 1$ to $E$}
    \State $h \gets e^{\theta}$
    \Comment{Recover positive shape parameter from log-space}
    
    \Statex
    \State \textit{Forward pass:} 
    compute edge functions for all layers $k$ 
    and neurons $m, n$:
    \Statex
    \State \quad $\psi_{k,m,n}(x) = 
    \displaystyle\sum_{j=1}^{K} 
    c_{k,m,n,j}\,\phi\!\left(\frac{|x - c_j|}{h}\right)$
    \Statex
    \State \quad then assemble the network output 
    via \eqref{Kan-out} to obtain $\hat{y}_i$ 
    for each training point $\mathbf{x}_i$
    
    \Statex
    \State \textit{Compute loss:}
    \State \quad $\mathcal{L} \gets 
    \dfrac{1}{N_{train}} 
    \sum_{i=1}^{N_{train}} 
    (y_i - \hat{y}_i)^2$
    \Comment{Mean squared error}
    
    \Statex
    \State \textit{Backpropagation:} update 
    $\theta$ and $\{c_{k,m,n,j}, w_{k,m,n}\}$ 
    via Adam with learning rate $\alpha$
    
    \Statex
    \If{$e \bmod 100 = 0$}
        \State Evaluate on $\mathcal{D}_{test}$; 
        record current $h = e^\theta$ 
        and validation loss
    \EndIf
\EndFor

\Statex
\State \textbf{Step 4: Surface Reconstruction}
\State Evaluate 
$\hat{y}_i = 
(\boldsymbol{\Psi}_{L-1} \circ \cdots \circ 
\boldsymbol{\Psi}_0)(\mathbf{x}_i)$
for each point $\mathbf{x}_i$ in the domain
to obtain $\hat{\mathbf{y}}$
\Comment{Composition as in \eqref{Kan-out}, 
using trained RBF edge functions; 
grid details in Section~\ref{sec:results}}

\Statex
\State \Return 
$\{\boldsymbol{\Psi}_k\}_{k=0}^{L-1}$, 
$\hat{\mathbf{y}}$

\end{algorithmic}
\end{algorithm}

\newpage
\section{Numerical Results} \label{sec:results}

In this section, we present the numerical results of the proposed adaptive RBF-KAN. The experiments are designed to evaluate whether adapting the shape parameter $h$ during training improves approximation 
accuracy over a fixed parameter, and to compare the proposed model against several baseline architectures including MLPs, the standard KAN, and existing KAN variants.

All the test functions are defined on the unit square domain $[0, 1]^2$. These functions were selected because they represent structurally distinct approximation challenges, including smooth surfaces, sharp discontinuities, high-frequency oscillations, 
and localized singularities, allowing a comprehensive evaluation of the model across different types of target functions.

\subsection{Target benchmark functions}

\textbf{Franke Function (Smooth):} This function is a standard benchmark for smooth surface reconstruction. It has two Gaussian peaks and a sharp hill and is infinitely differentiable ($C^{\infty}$):
\begin{equation*}
    \begin{aligned}
        f_1(x, y) & =  0.75 \exp\left(-\frac{(9x-2)^2 + (9y-2)^2}{4}\right) + 0.75 \exp\left(-\frac{(9x+1)^2}{49} - \frac{9y+1}{10}\right) \\
        & \,\,\,\,\, + 0.5 \exp\left(-\frac{(9x-7)^2 + (9y-3)^2}{4}\right) - 0.2 \exp\left(-(9x-4)^2 - (9y-7)^2\right).
    \end{aligned}
\end{equation*}

\textbf{Discontinuous Function (Sharp Jump):} This function has a circular step discontinuity. It is used to test if the model can learn sharp transitions without making oscillations (i.e. Gibbs phenomena) around the jump:
\begin{equation*}
    f_2(x, y) = \begin{cases} 
    1 & \text{if } \sqrt{x^2 + y^2} \geq 0.5, \\
    0 & \text{otherwise}.
    \end{cases}
\end{equation*}

\textbf{High-Frequency Function (Oscillatory):} 
This function tests the spatial resolution of 
the fixed RBF center grid 
$\{c_j\}_{j=1}^{K}$ used in each edge function 
(see Algorithm~\ref{alg3}). It shows if the model can capture rapid, repeating patterns:
\begin{equation*}
    f_3(x, y) = \sin(25x) \cos(25y).
\end{equation*}

\textbf{Singularity Function (Local Peak):} This function has a peak with a very high gradient that approaches a singularity point at $(0.5, 0.5)$. It is used to check the approximation accuracy in areas where the data is not very smooth:
\begin{equation*}
    f_4(x, y) = \frac{1}{\sqrt{(x-0.5)^2 + (y-0.5)^2} + 0.1}.
\end{equation*}

\subsection{Experimental setup}

For a completely fair comparison between our adaptive FastKAN, the standard KAN, variants of KAN and the MLP, we trained all the networks under the exact same conditions. None of the models is discriminated by either providing more data, epochs or learning rate. All models were trained using FP64 precision on a Linux-based workstation equipped with an NVIDIA H100 GPU and 256GB of RAM.

Table \ref{tab:fair_comparison_setup} lists all the common training parameters fixed during training of all the networks.

\begin{table}[h]
\centering
\caption{Experimental parameters for fair comparison.}
\label{tab:fair_comparison_setup}
\begin{tabular}{ll}
\hline
\textbf{Parameter} & \textbf{Value} \\
\hline
Total Data Samples ($N$) & 2000 points \\
Data Split & 80\% Training, 20\% Testing \\
Input Dimension & 2 ($x$ and $y$ coordinates) \\
Output Dimension & 1 ($z$ target value) \\
Optimizer & Adam \\
Learning Rate ($\alpha$) & 1e-02 \\
Total Training Epochs (E) & 2000 \\
\hline
\end{tabular}
\end{table}

Since all networks were trained with the same data 
points, optimizer, learning rate, and number 
of epochs, any differences in the final $L_2$ error and execution time reflect 
differences in model architecture alone. We 
also tested MLP with up to $20{,}000$ epochs 
and a target loss of $10^{-3}$, but it did not 
converge to this threshold; the results at 
$2{,}000$ epochs were therefore retained for 
all models to ensure a fair and consistent 
comparison.

Table~\ref{tab:LOOCV_results} presents the 
detailed results of the adaptive RBF-KAN using 
different kernels. The original FastKAN model 
uses the GA kernel with a fixed shape parameter. 
Our experiments reveal that a fixed shape 
parameter is not an optimal choice for all 
problem types. While the GA kernel works well 
for the smooth function $f_1$, it struggles with 
the discontinuous jump in $f_2$, where the M2 
kernel provides a significantly lower error. For 
the oscillatory function $f_3$, GA completely 
fails whereas W2 captures the pattern accurately. 
For the singularity function $f_4$, W6 reduces 
the error by approximately 52\% over the GA 
baseline. These results confirm that adaptive 
kernel selection with LOOCV-based initialization 
is more robust than the fixed GA approach of 
standard FastKAN.

\begin{figure}[htbp]
    \centering
    \begin{subfigure}[b]{0.48\textwidth}
        \centering
        \includegraphics[width=\textwidth]{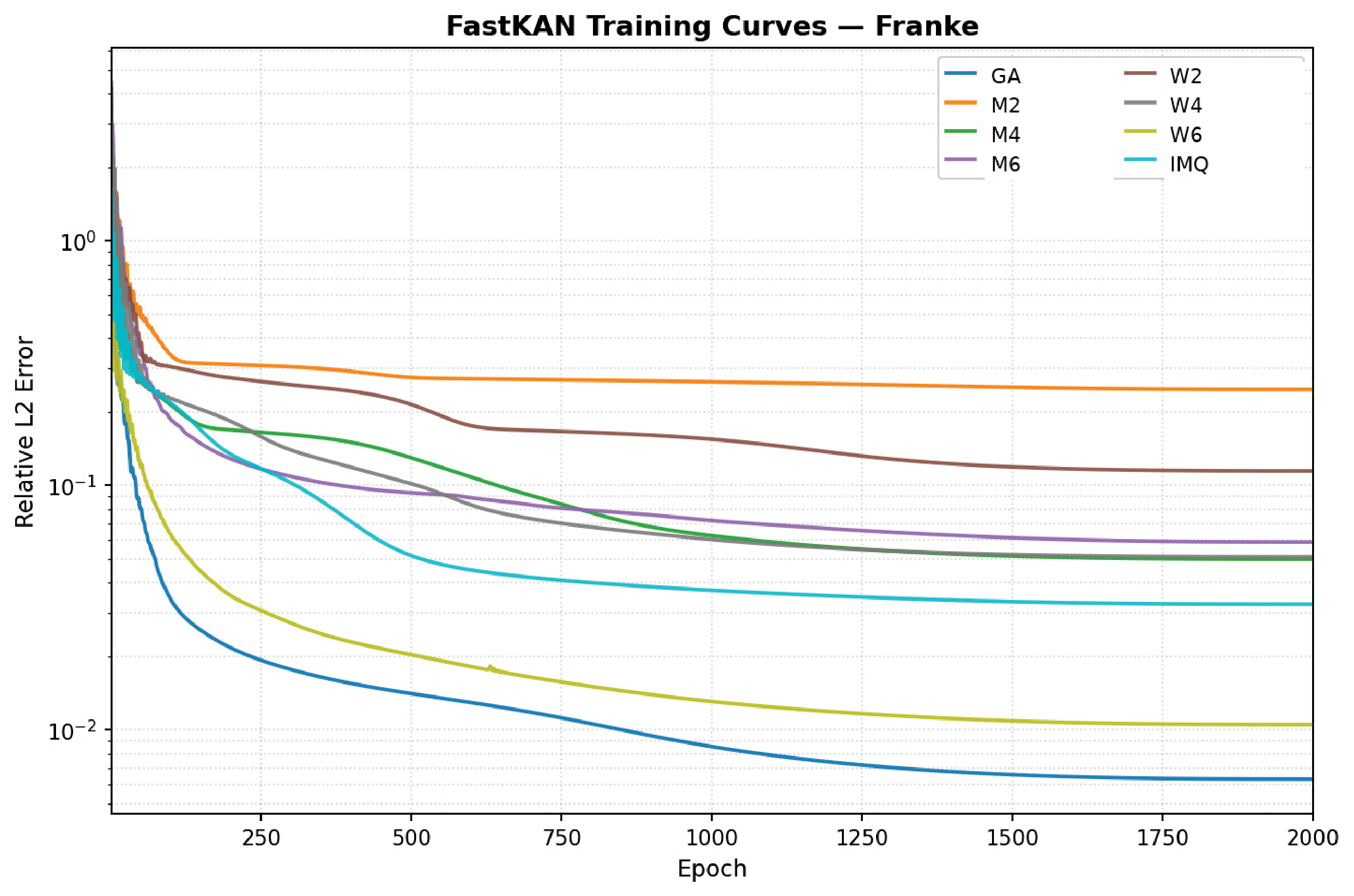}
        \caption{$f_1$}
        \label{fig:franke}
    \end{subfigure}
    \hfill
    \begin{subfigure}[b]{0.48\textwidth}
        \centering
        \includegraphics[width=\textwidth]{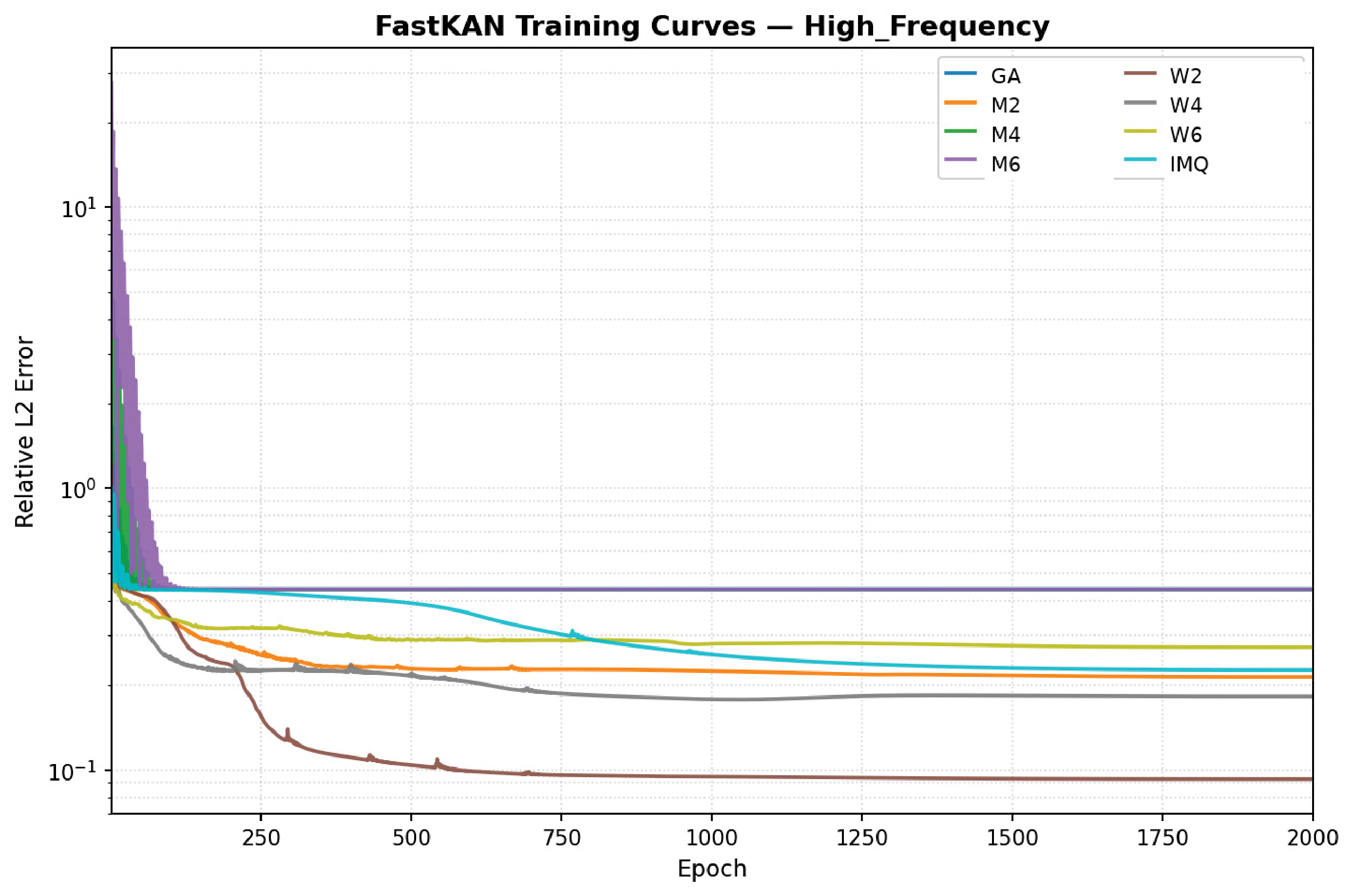}
        \caption{$f_3$}
        \label{fig:highfreq}
    \end{subfigure}

    \vspace{0.5em}

    \begin{subfigure}[b]{0.48\textwidth}
        \centering
        \includegraphics[width=\textwidth]{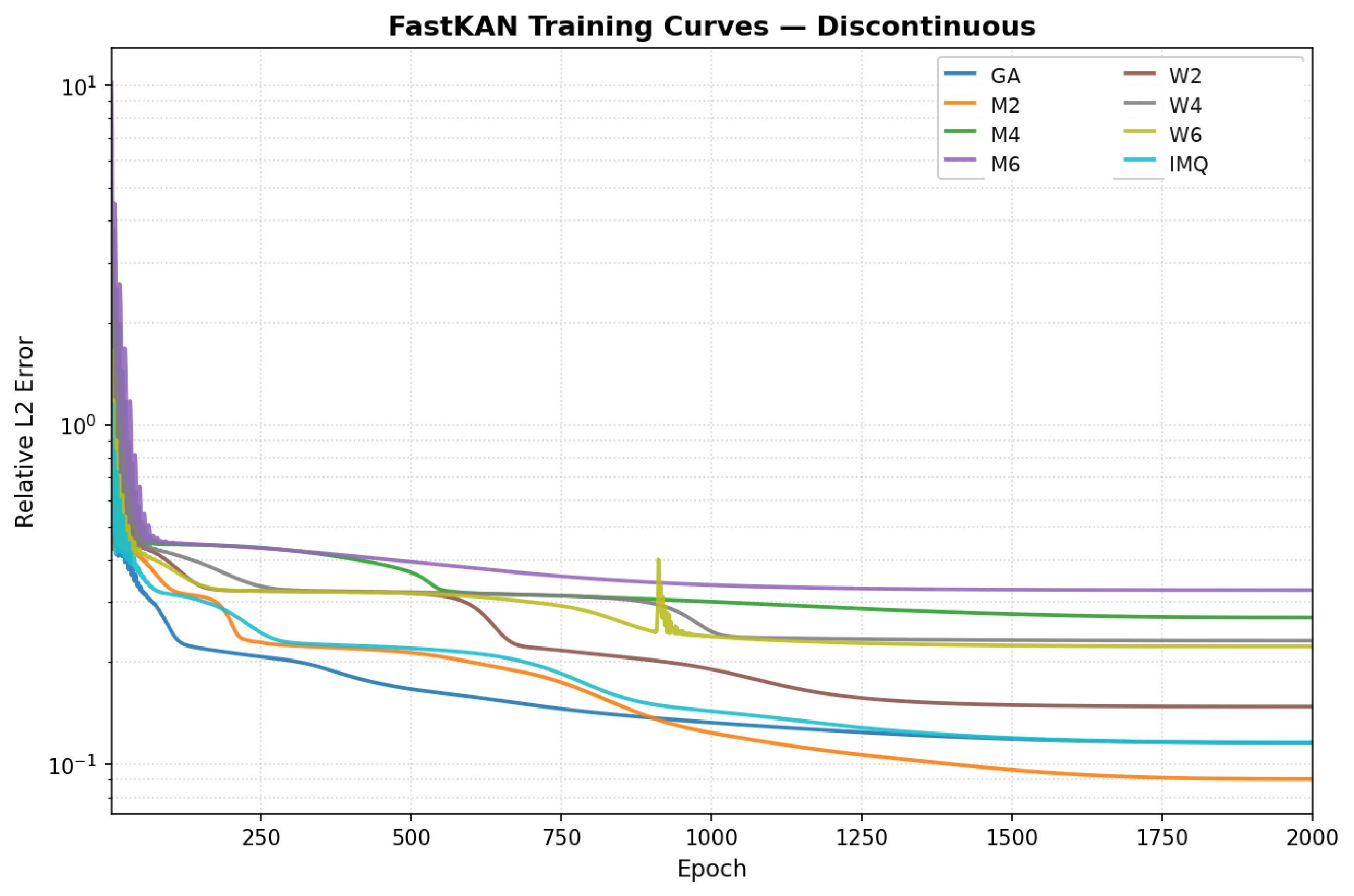}
        \caption{$f_2$}
        \label{fig:discontinuous}
    \end{subfigure}
    \hfill
    \begin{subfigure}[b]{0.48\textwidth}
        \centering
        \includegraphics[width=\textwidth]{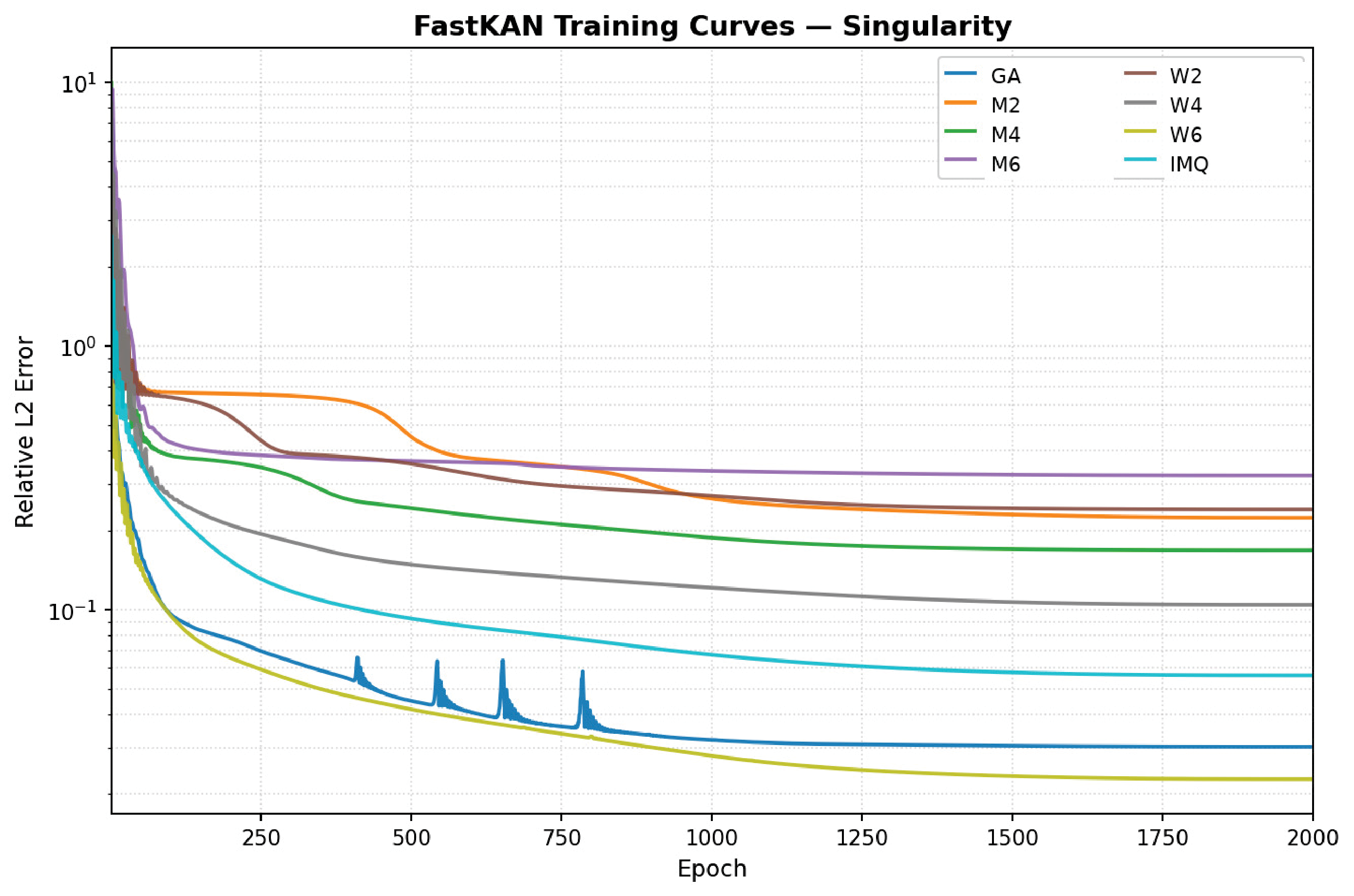}
        \caption{$f_4$}
        \label{fig:singularity}
    \end{subfigure}

    \caption{Training curves (Relative $L_2$ Error vs.\ Epoch) for adaptive FastKAN with 
    different RBF kernels across four benchmark functions.}
    \label{fig:training_curves}
\end{figure}

\begin{table}[htbp]
\centering
\caption{Experimental Results for adaptive FastKAN.}
\label{tab:LOOCV_results}
\small
\begin{tabular}{l l c c c r}
\toprule
\textbf{Function} & \textbf{Kernel} & \textbf{Initial $h_{opt}$ (LOOCV)} & \textbf{Final $h$} & \textbf{Rel.\ $L_2$ Error} & \textbf{Time (s)} \\
\midrule
\multirow{8}{*}{\textbf{$f_1$}} 
& GA   & 0.18 & 0.19 & 4.07e-03 & 3.76 \\
& M2   & 0.34 & 0.28 & 1.84e-01 & 3.77 \\
& M4   & 0.18 & 0.16 & 1.45e-02 & 4.34 \\
& M6   & 0.26 & 0.25 & 1.43e-01 & 4.69 \\
& W2   & 2.78 & 1.84 & 5.16e-02 & 4.29 \\
& W4   & 1.08 & 1.02 & 1.42e-02 & 4.72 \\
& W6   & 1.62 & 1.16 & 2.06e-02 & 5.13 \\
& IMQ  & 0.38 & 0.19 & 2.83e-01 & 3.65 \\
\midrule
\multirow{8}{*}{\textbf{$f_2$}} 
&GA    & 7.12 & 0.21 & 1.19e-01 & 3.67 \\
&M2   & 1.72 & 0.16 & 3.94e-02 & 3.74 \\
&M4   & 11.21 & 0.23 & 2.84e-01 & 4.34 \\
&M6   & 3.37 & 2.70 & 3.13e-01 & 4.74 \\
&W2 & 6.10 & 0.98 & 4.32e-02 & 4.16 \\
&W4 & 5.80 & 3.14 & 1.03e-01 & 4.67 \\
&W6 & 17.73 & 11.15 & 1.17e-01 & 5.07 \\
&IMQ  & 1.93 & 0.18 & 5.70e-02 & 3.90 \\
\midrule
\multirow{8}{*}{\textbf{$f_3$}} 
&GA    & 3.06 & 2.49 & 4.56e-01 & 3.86 \\
&M2   & 0.04 & 0.09 & 7.37e-02 & 3.99 \\
&M4   & 14.84 & 14.48 & 4.56e-01 & 4.52 \\
&M6   & 20.00 & 21.47 & 4.57e-01 & 5.13 \\
&W2 & 0.25 & 0.46 & 6.24e-02 & 4.77 \\
&W4 & 0.28 & 0.50 & 7.09e-02 & 6.02 \\
&W6 & 0.35 & 0.49 & 8.68e-02 & 6.58 \\
&IMQ  & 4.65 & 4.30 & 4.56e-01 & 3.93 \\
\midrule
\multirow{8}{*}{\textbf{$f_4$}} 
&GA    & 0.08 & 0.17 & 4.02e-02 & 3.80 \\
&M2   & 0.05 & 0.09 & 4.22e-02 & 3.84 \\
&M4   & 0.04 & 0.05 & 4.42e-02 & 4.16 \\
&M6   & 0.03 & 0.04 & 1.67e-01 & 4.50 \\
&W2 & 0.32 & 0.43 & 3.13e-02 & 4.24 \\
&W4 & 0.49 & 0.48 & 2.63e-02 & 4.74 \\
&W6 & 0.39 & 0.50 & 1.92e-02 & 5.12 \\
&IMQ  & 0.07 & 0.09 & 4.85e-02 & 3.77 \\
\bottomrule
\end{tabular}
\end{table}
Figure~\ref{fig:training_curves} shows the 
training curves for all eight kernels across 
the four benchmark functions. The convergence 
behavior varies significantly between kernels, 
particularly for $f_2$ and $f_3$, where 
compact-support kernels converge to lower 
errors than global kernels such as GA and IMQ.

\begin{figure}[htbp]
    \centering
    \includegraphics[width=0.8\textwidth]{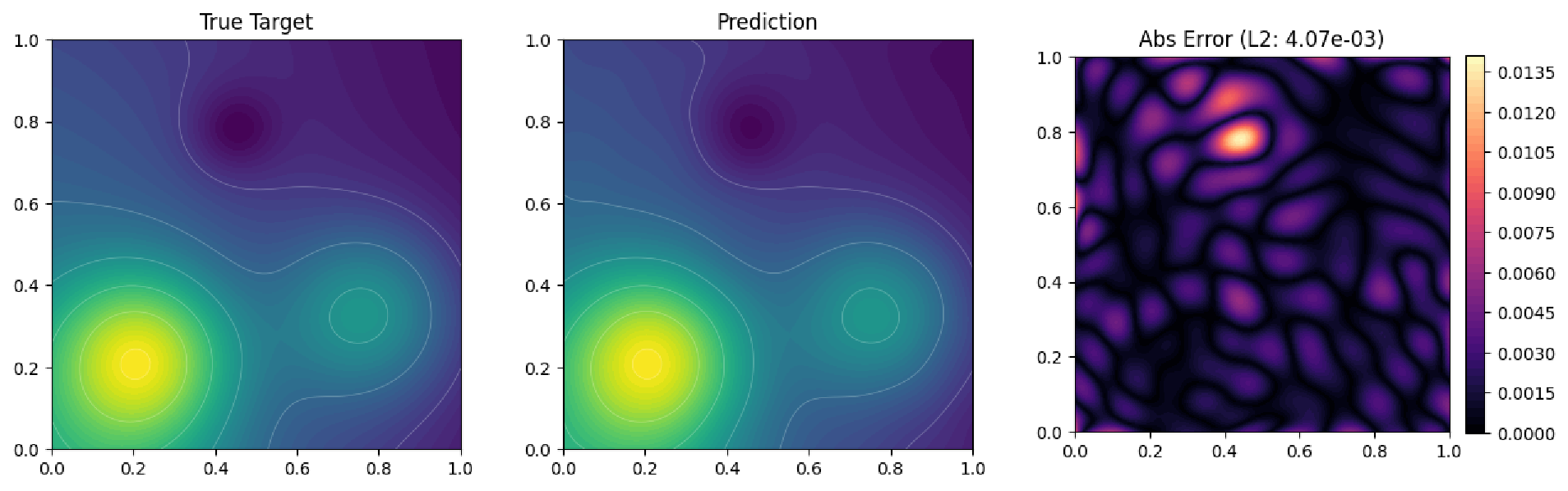}
    \caption{Reconstruction of $f_1$ with GA kernel. GA works best here because the function is very smooth ($L_2$ error: 4.07e-3).}
    \label{fig:franke_best}
\end{figure}

\begin{figure}[htbp]
    \centering
    \begin{subfigure}{0.8\textwidth}
        \centering
        \includegraphics[width=\textwidth]{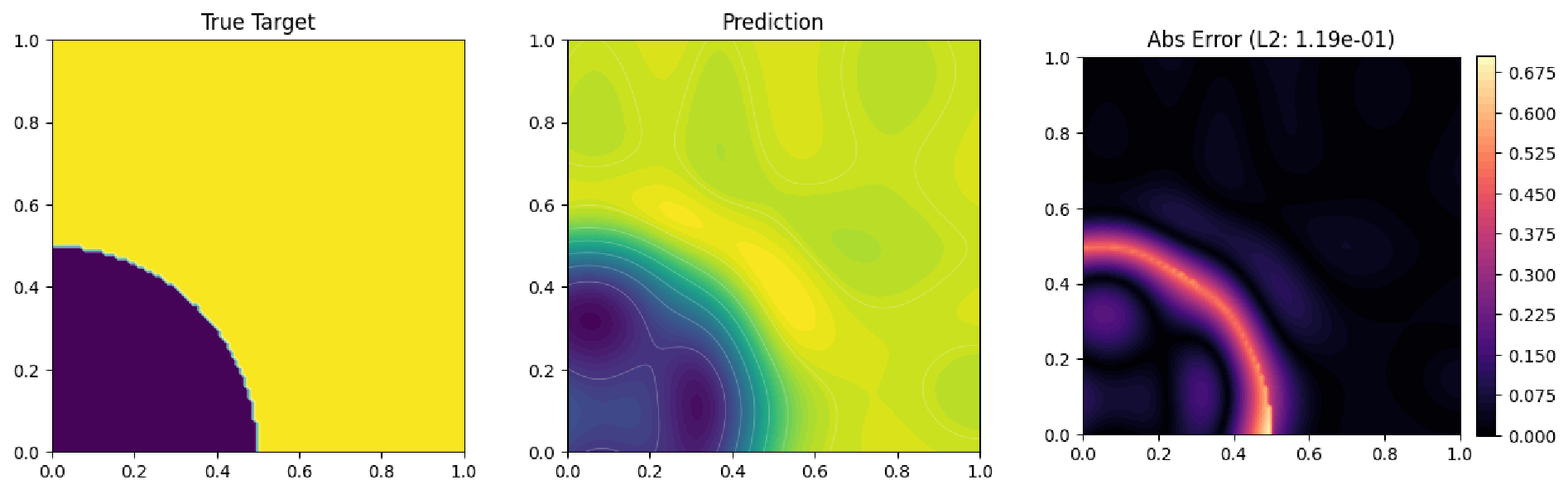}
        \caption{Baseline: GA ($L_2$: 1.19e-1)}
    \end{subfigure}\hfill
    \begin{subfigure}{0.8\textwidth}
        \centering
        \includegraphics[width=\textwidth]{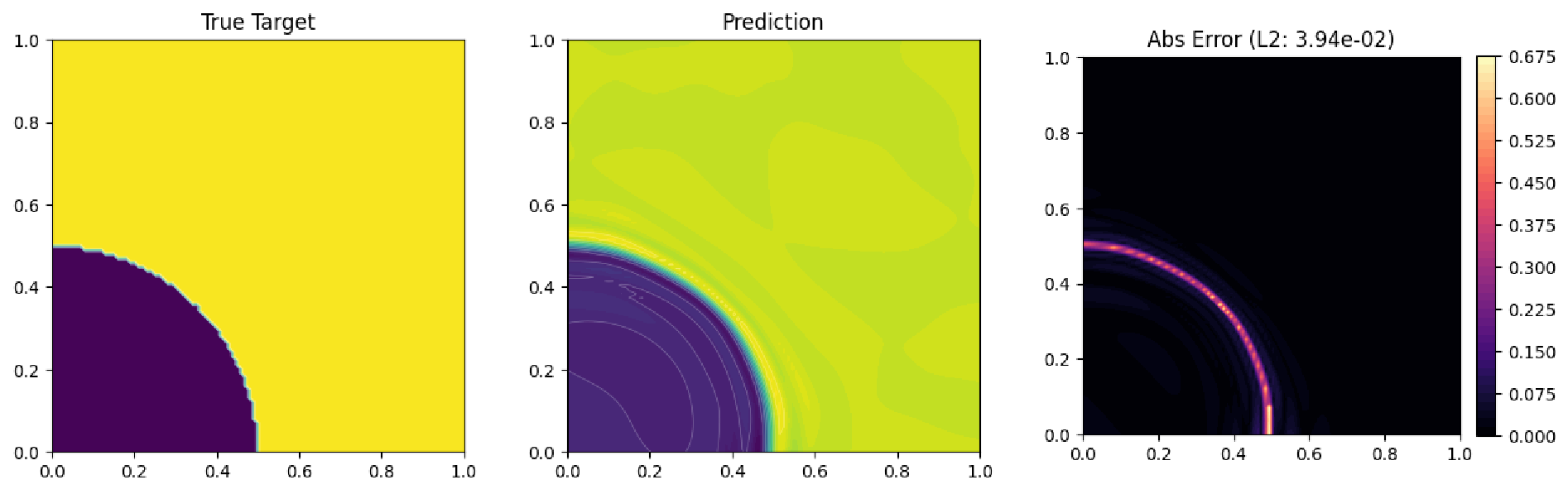}
        \caption{Best: M2 ($L_2$: 3.94e-2)}
    \end{subfigure}
    \caption{Reconstruction of $f_2$. M2 captures the sharp jump much better than GA. }
    \label{fig:comp_discontinuous}
\end{figure}

\begin{figure}[htbp]
    \centering
    \begin{subfigure}{0.8\textwidth}
        \centering
        \includegraphics[width=\textwidth]{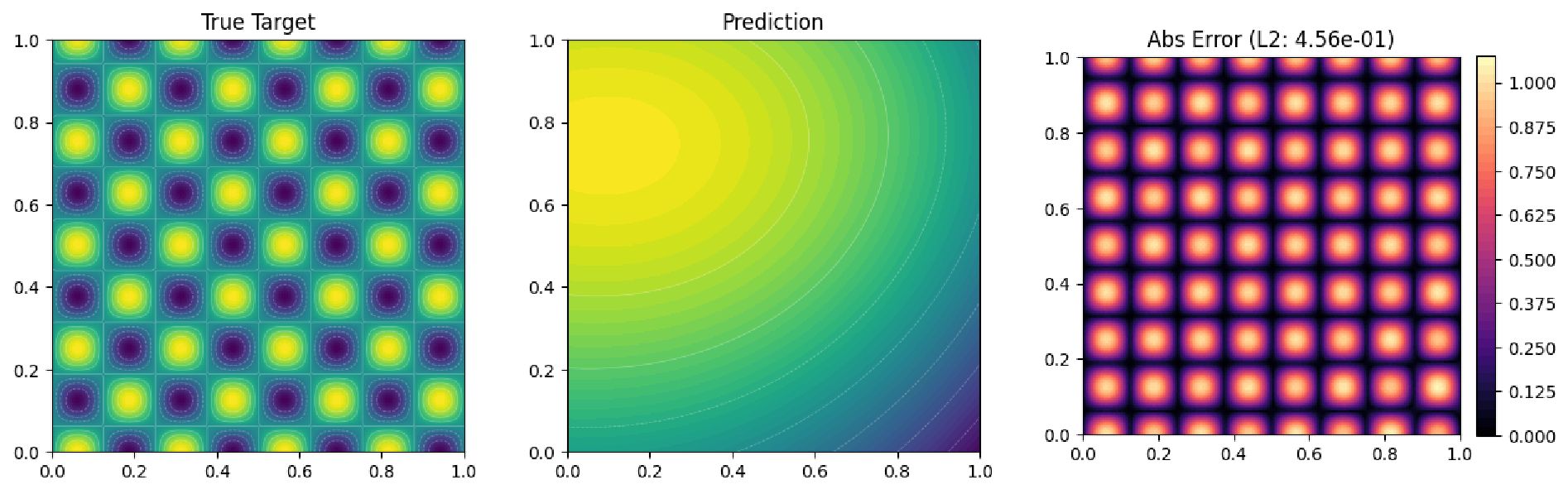}
        \caption{Baseline: GA ($L_2$: 4.56e-1)}
    \end{subfigure}\hfill
    \begin{subfigure}{0.8\textwidth}
        \centering
        \includegraphics[width=\textwidth]{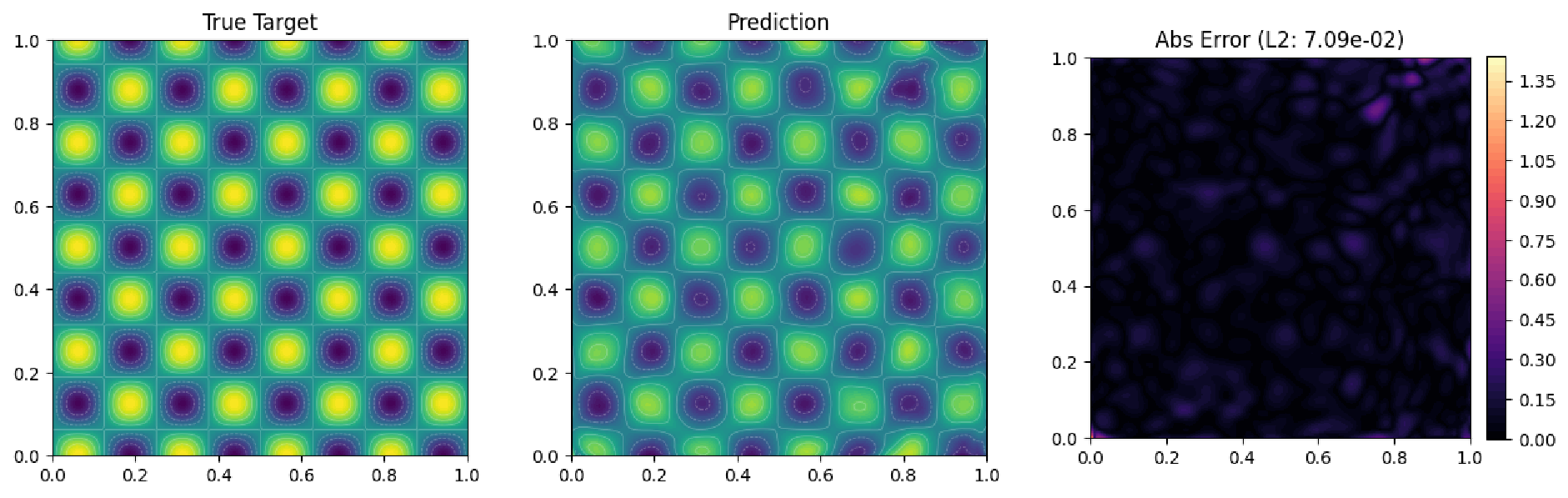}
        \caption{Best: W2 ($L_2$: 6.24e-2)}
    \end{subfigure}
    \caption{Reconstruction of $f_3$. GA completely misses the waves, but W2 rebuilds the pattern well. }
    \label{fig:comp_high_freq}
\end{figure}

\begin{figure}[htbp]
    \centering
    \begin{subfigure}{0.8\textwidth}
        \centering
        \includegraphics[width=\textwidth]{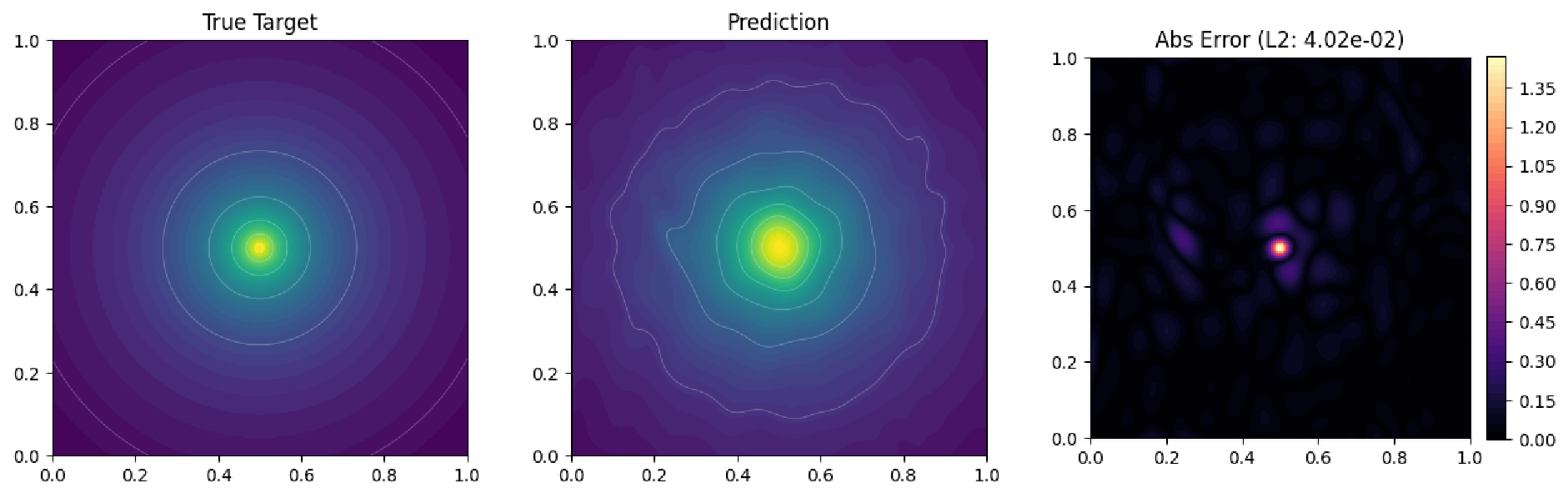}
        \caption{Baseline: GA ($L_2$: 4.02e-2)}
    \end{subfigure}\hfill
    \begin{subfigure}{0.8\textwidth}
        \centering
        \includegraphics[width=\textwidth]{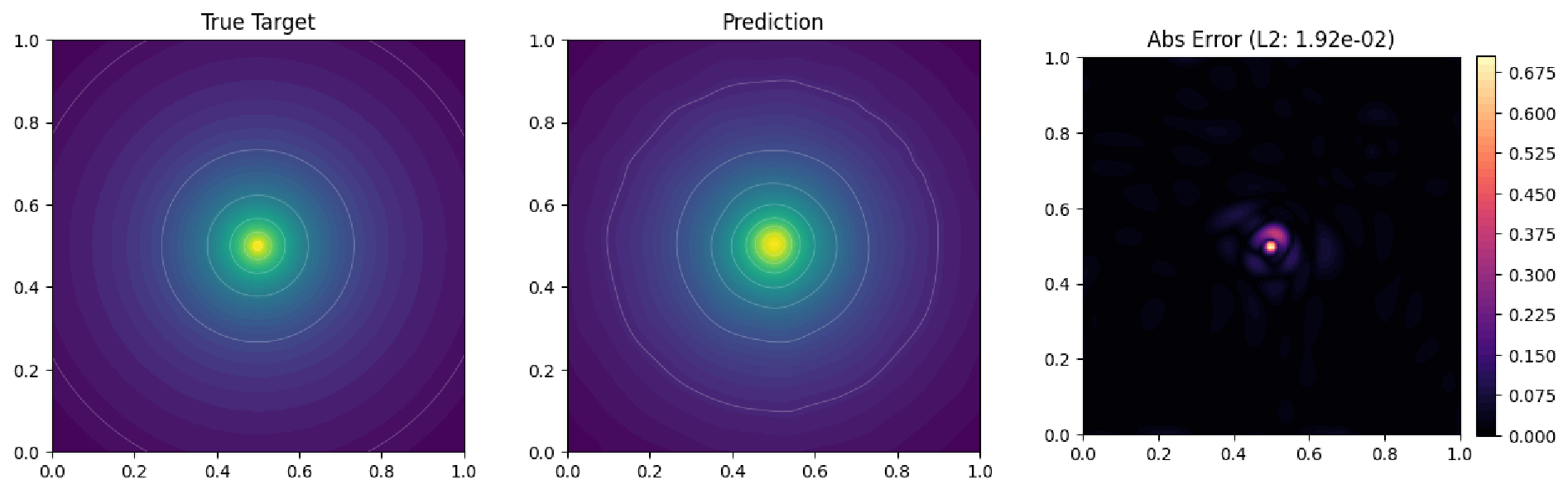}
        \caption{Best: W6 ($L_2$: 1.92e-2)}
    \end{subfigure}
    \caption{Reconstruction of $f_4$. W6 handles the sharp peak better, giving 52\% less error than GA. }
    \label{fig:comp_singularity}
\end{figure}

The contour plots confirm these observations 
visually. For $f_1$ (Figure~\ref{fig:franke_best}), the GA kernel 
produces an accurate reconstruction, consistent 
with its strong performance on infinitely smooth 
functions. For $f_2$ 
(Figure~\ref{fig:comp_discontinuous}), the GA 
kernel blurs the sharp circular boundary while 
the M2 kernel resolves it accurately. For $f_3$ 
(Figure~\ref{fig:comp_high_freq}), the GA kernel 
fails to capture the oscillatory pattern whereas 
the W2 kernel reconstructs it accurately. For 
$f_4$ (Figure~\ref{fig:comp_singularity}), the 
W6 kernel resolves the sharp central peak more 
accurately than GA, consistent with the error 
reduction reported above. These visual results 
confirm that adaptive kernel selection with 
LOOCV-based shape parameter initialization 
provides consistent improvements over the fixed 
GA approach of standard FastKAN.

Table \ref{tab:final_comparison} presents the overall comparative analysis of our adaptive FastKAN and other KAN variants, i.e. EfficientKAN, Chebyshev KAN, KAF, standard KAN, and MLP.
From the results, it is clear that using a fixed shape parameter of value 0.5714, as suggested in \cite{li24}, does not perform well in terms of accuracy. On the other hand, a significant error drop in term of relative $L_2$  norm could be observed in the case of our proposed model with adaptive shape parameter and choice of best kernel. As an example, for $f_3$ the error improves from 4.54e-01 to 6.24e-02 just by adapting the shape parameter and using the W2 kernel. 

From Table \ref{tab:final_comparison} we can observe that in most cases the standard KAN gives accurate reconstruction of the function but it is computationally expensive. On the other hand, though the MLP is robust, it fails to reach a certain level of accuracy on complex patterns like $f_3$ and $f_4$. While all other KAN variants perform well in some cases, our adaptive FastKAN provides an overall balance between accuracy and efficiency in all cases. It achieves competitive accuracy across all functions considered while maintaining training times comparable to FastKAN.
\begin{table}[htbp]
\centering
\caption{Comparison of various KAN versions and MLP.}
\label{tab:final_comparison}
\small
\begin{tabular}{l l c c S[table-format=1.4e-2] S[table-format=2.2]}
\toprule
\textbf{Function} & \textbf{Method} & \textbf{Shape par. ($h$)} & \textbf{Architecture} & {\textbf{Rel. $L_2$ Error}} & {\textbf{Time (s)}} \\
\midrule
\multirow{7}{*}{\textbf{$f_1$}} &  FastKAN (GA) & 0.5714 (fixed) & [2, 8, 1] & 4.08e-02 & 2.47 \\
 & \textbf{Adaptive FastKAN (GA)} & 0.19 & [2, 8, 1]  & 4.07e-03 & 3.76 \\
 & Efficient KAN & - & [2, 8, 1] & 5.93e-03 & 7.38 \\
 & Chebyshev KAN & - & [2, 8, 1] & 6.96e-03 & 2.57 \\
 & KAF & - & [2, 5, 5, 1] & 3.92e-02 & 2.00 \\
 & Standard KAN & - & [2, 5, 5, 1] & 4.21e-02 & 62.92 \\
 & MLP & - & [2, 128, 128, 128, 1]  & 3.04e-02 & 1.88 \\
\midrule
\multirow{7}{*}{\textbf{$f_2$}} &  FastKAN (GA) & 0.5714 (fixed) & [2, 8, 1] & 4.98e-02 & 2.28 \\
 & \textbf{Adaptive FastKAN (M2)} & 0.16 & [2, 8, 1] & 3.94e-02 & 3.74 \\
 & Efficient KAN & - & [2, 8, 1] & 1.22e-02 & 7.23 \\
 & Chebyshev KAN & - & [2, 8, 1] & 6.09e-02 & 2.48 \\
 & KAF & - & [2, 5, 5, 1] & 4.52e-02 & 1.86 \\
 & Standard KAN & - & [2, 5, 5, 1] & 5.64e-02 & 61.87 \\
 & MLP & - & [2, 128, 128, 128, 1] & 5.57e-02 & 1.86 \\
\midrule
\multirow{7}{*}{\textbf{$f_3$}} & FastKAN (GA) & 0.5714 (fixed) & [2, 16, 1] & 4.54e-01 & 2.42 \\
 & \textbf{Adaptive FastKAN (W2)} & 0.46 & [2, 16, 1] & 6.24e-02 & 4.77 \\
 & Efficient KAN & - & [2, 16, 1] & 1.37e-01 & 7.30 \\
 & Chebyshev KAN & - & [2, 16, 1] & 6.47e-02 & 3.78 \\
 & KAF & - & [2, 5, 5, 1] & 4.49e-01 & 1.94 \\
 & Standard KAN & - & [2, 5, 5, 1] & 1.65e-01 & 62.13 \\
 & MLP & - & [2, 128, 128, 128, 1] & 4.34e-01 & 1.91 \\
\midrule
\multirow{7}{*}{\textbf{$f_4$}} & FastKAN (GA) & 0.5714 (fixed) & [2, 8, 1] & 4.25e-02 & 2.29 \\
 & \textbf{Adaptive FastKAN (W6)} & 0.50 & [2, 8, 1] & 1.92e-02 & 5.12 \\
 & Efficient KAN & - & [2, 8, 1] & 1.79e-02 & 7.27 \\
 & Chebyshev KAN & - & [2, 8, 1] & 6.27e-02 & 2.48 \\
 & KAF & - & [2, 5, 5, 1] & 1.57e-01 & 1.85 \\
 & Standard KAN & - & [2, 5, 5, 1] & 8.53e-02 & 62.08 \\
 & MLP & - & [2, 128, 128, 128, 1] & 1.07e-01 & 1.79 \\
\bottomrule
\end{tabular}
\end{table}

\newpage

\section{Conclusion} \label{sec:conclusion}

In this work, we investigated the use of KANs and their variants for function approximation and introduced an adaptive RBF-based extension of the FastKAN architecture. The proposed adaptive RBF-KAN model combines RBF representations with a data-driven initialization of the kernel shape parameter using LOOCV, followed by further refinement during network training.

To the best of our knowledge, this work represents the first attempt to integrate LOOCV-based kernel scale estimation with the training of deep KAN architectures. In addition, we introduced several RBFs--including Matérn and Wendland kernels--into the KAN framework for the first time, extending the original FastKAN formulation which relies solely on the GA kernel.

The proposed method was evaluated on four benchmark functions defined on the unit square, representing different approximation challenges such as smooth behavior, discontinuities, oscillatory patterns, and localized singularities.

There are several key observations that are highlighted in the paper. The results show that MLP networks are computationally efficient but suffer from accuracy limitations and tend to over-smooth while dealing with complex structures, particularly in the presence of high-frequency patterns. On the other hand, the standard KAN is observed to be capable of providing higher accuracy but is computationally expensive due to the spline basis functions. The proposed adaptive RBF-KAN proves to be a practical solution as it maintains the computational efficiency of FastKAN and also improves the flexibility of the network through the choice of an appropriate kernel for specific problem and an adaptive shape parameter.

Furthermore, the results demonstrate that the selection of appropriate RBF kernel plays an important role in approximation performance. For smooth functions the GA kernels outperformed, whereas Matérn kernels provide improved accuracy for functions with a circular discontinuity. Wendland kernels exhibit stable convergence and efficient performance for oscillatory functions and localized singularities. 

Overall, the results indicate that combining LOOCV-based initialization with adaptive kernel learning provides a flexible strategy for improving RBF-based KAN models across different types of data. Future work will investigate the application of this framework to higher-dimensional problems, noisy real-world datasets, and scientific computing applications such as the solution of partial differential equations.


\section*{Acknowledgments}

This work has been supported by the INdAM Research group GNCS, the GFI 2025 Project, and the 2024 Project \lq\lq Numerical Analysis and Modelling\rq\rq\ funded by the Department of Mathematics \lq\lq Giuseppe Peano\rq\rq\ of the University of Torino. This research has been accomplished within the RITA \lq\lq Research ITalian network on Approximation\rq\rq, the UMI Group TAA \lq\lq Approximation Theory and Applications\rq\rq, and the SIMAI Activity Group ANA$\&$A \lq\lq Numerical and Analytical Approximation of Data and Functions with Applications\rq\rq.







\end{document}